\documentclass[10pt,twocolumn,letterpaper]{article}

\usepackage[numbers,square]{natbib}
\usepackage{mathptmx}
\usepackage{graphicx}
\usepackage{amsmath}
\usepackage{amssymb}
\usepackage{multirow}
\usepackage{tabularx}
\usepackage{booktabs}
\usepackage{standalone}
\usepackage{xcolor}
\usepackage{tikz}
\usetikzlibrary{snakes,arrows,shapes}
\usepackage{pgfplots}
\pgfplotsset{compat = 1.3}
\usepackage{wrapfig}

\usepackage{amssymb,amsmath}
\usepackage{algorithm}
\usepackage{algorithmic}
\usepackage{graphicx}
\usepackage{multirow}
\usepackage{array}
\usepackage{graphicx}
\usepackage{caption}
\usepackage{subcaption}

\usepackage{tabularx}
\newcolumntype{L}{>{\raggedright\arraybackslash}X}\newcolumntype{C}{>{\centering\arraybackslash}X}\newcolumntype{R}{>{\raggedleft\arraybackslash}X}

\usepackage{cvpr}
\def\cvprPaperID{}

\newcommand{\tick}{\checkmark}

\usepackage[breaklinks=true,bookmarks=false]{hyperref}

\cvprfinalcopy % *** Uncomment this line for the final submission

\def\cvprPaperID{****} % *** Enter the CVPR Paper ID here

\ifcvprfinal\fi
\begin{document}

%%%%%%%%% TITLE
\title{Truly Multi-modal YouTube-8M Video Classification with Video, Audio, and Text}

\author{
     Zhe Wang\footnotemark[1]
   \\{\tt\small mark.wangzhe@gmail.com}
\and Kingsley Kuan\footnotemark[1]
   \\{\tt\small kingsley.kuan@gmail.com}
\and Mathieu Ravaut\footnotemark[1]
   \\{\tt\small mathieu.ravaut@student.ecp.fr}
\and Gaurav Manek\footnotemark[1]
   \\{\tt\small manekgm@i2r.a-star.edu.sg}
\and Sibo Song\thanks{Joint first-authorship.}
   \\{\tt\small sibo\_song@mymail.sutd.edu.sg}
\and Yuan Fang
   \\{\tt\small yfang@i2r.a-star.edu.sg}
\and Seokhwan Kim
   \\{\tt\small kims@i2r.a-star.edu.sg}
\and Nancy F. Chen
   \\{\tt\small nfychen@i2r.a-star.edu.sg}
\and Luis Fernando D'Haro
   \\{\tt\small luisdhe@i2r.a-star.edu.sg}
\and Luu Anh Tuan
   \\{\tt\small at.luu@i2r.a-star.edu.sg}
\and Hongyuan Zhu
   \\{\tt\small zhuh@i2r.a-star.edu.sg}
\and Zeng Zeng
   \\{\tt\small zengz@i2r.a-star.edu.sg}
\and Ngai Man Cheung
   \\{\tt\small ngaiman\_cheung@sutd.edu.sg}
\and Georgios Piliouras
   \\{\tt\small georgios@sutd.edu.sg}
\and Jie Lin
   \\{\tt\small lin-j@i2r.a-star.edu.sg}
\and Vijay Chandrasekhar
   \\{\tt\small vijay@i2r.a-star.edu.sg}
}

\maketitle
%\thispagestyle{empty}

%%%%%%%%% ABSTRACT
\begin{abstract}
The \texttt{YouTube-8M} video classification challenge requires teams to classify 0.7 million videos into one or more of 4,716 classes.
In this Kaggle competition, we placed in the top 3\% out of 650 participants using released video and audio features.

Beyond that, we extend the original competition by including text information in the classification, making this a truly multi-modal approach with vision, audio and text.
The newly introduced text data is termed as YouTube-8M-Text.
We present a classification framework for the joint use of text, visual and audio features, and conduct an extensive set of experiments to quantify the benefit that this additional mode brings.
The inclusion of text yields state-of-the-art results, e.g. 86.7\% GAP on the YouTube-8M-Text validation dataset.
\end{abstract}

\begin{figure*}[t]
        \centering
        \includegraphics[width=0.95\textwidth]{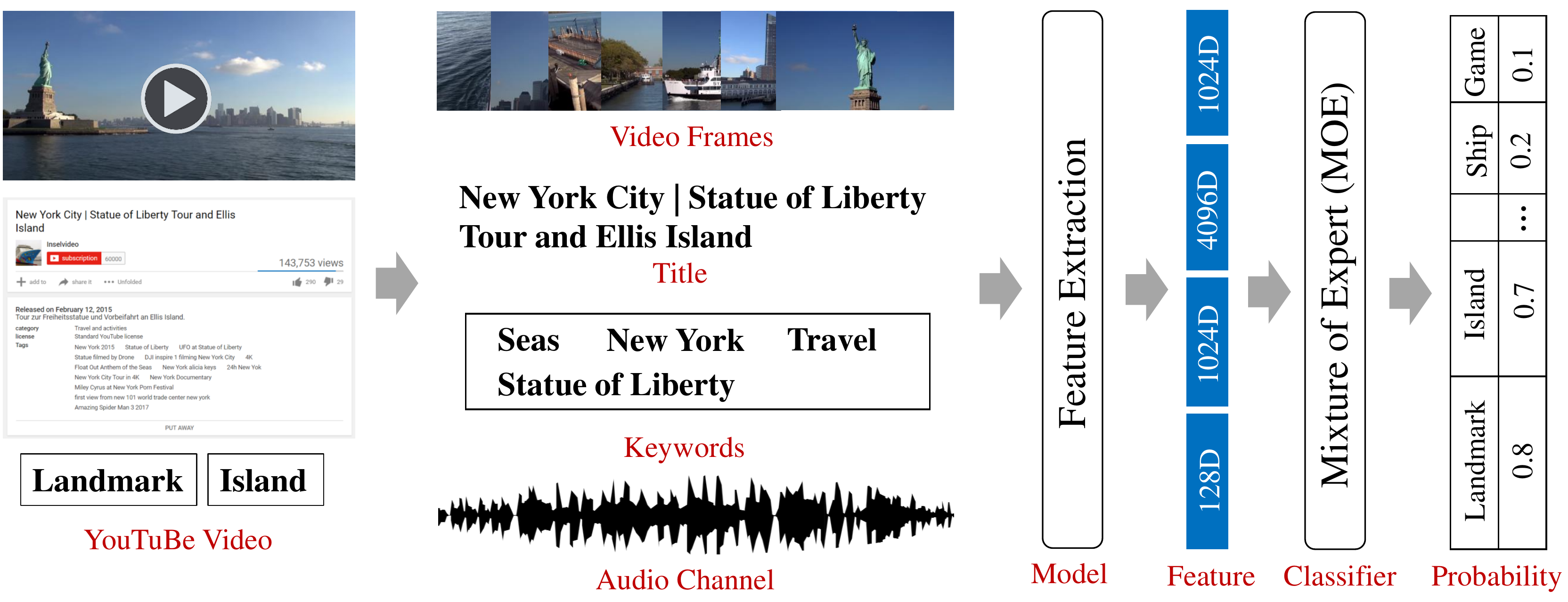}
        \caption{Overall framework for Youtube8M video classification with multi-modal fusion of video-level features, including visual, text and audio.}
        \label{fig:architecture_kaggle}
\end{figure*}

%%%%%%%%% BODY TEXT
\section{Introduction}

Video classification has long been an open problem of considerable academic interest.
With latest advances in neural networks and deep learning, new work has broken records on many key datasets. However, neural networks require large amounts of data for effective training.
Large-scale video data is something that only a few private companies have had access to until the recent \texttt{YouTube-8M} Kaggle competition;
then next largest video data set is the \texttt{Sports-1M} dataset~\cite{karpathy2014large} with 1.2 million videos over 480 classes.
This competition has spurred new interest in the video classification problem by providing a public dataset around which participants can rally.

This dataset presents two major challenges: diversity and class imbalance. As the members of the dataset were selected from all videos across YouTube,
they cover many different possible topics (music, politics, etc.), styles (CGI, surrealist, documentary, etc.), and formats (single-presenter, conversation, action, etc.).
Such diverse classes cannot necessarily be easily separated with low-level video and audio features alone:
for example, the difference between political satire and politics is not obvious, sometimes even to humans.

Further, an examination of the dataset reveals that there is a significant class imbalance, with only 10 labels accounting for over half of all label appearances in the dataset.
A large portion of classes has a very small number of examples, making it even difficult to bridge the so-called ``semantic gap'' between highly abstract labels and low-level features.

To address these problems, we make the following contributions in this work:
\begin{enumerate}
    \item The presence of an additional mode of data -- text -- can greatly improve classification performance by providing semantic information regarding the video.
    The surrounding text (e.g. titles, keywords) can disambiguate between videos that appear similar but require deep understanding to differentiate.
    By narrowing the semantic gap, we can potentially learn better representations and greatly improve performance on classes with very few examples.

    \item We propose a multi-model classification framework jointly modeling visual, audio and text data, making this a truly multi-modal approach.

    \item We conduct an extensive set of experiments to validate the effectiveness of text cues for the YouTube-8M video data set.
    Results show that the use of text significantly improves our solution on the Kaggle competition.
\end{enumerate}

Finally, we will release the \textbf{YouTube-8M-Text} dataset, the learned text features and text models to facilitate further research in YouTube-8M challenge.
The source codes, raw text data and tfrecord files of text features are available on \url{https://github.com/hrx2010/YouTube8m-Text}.

\section{Framework Overview}
We present the classification pipeline with multi-model video-level features in Fig.~\ref{fig:architecture_kaggle}.
We examine the performance improvement by concatenating text features with video and audio features, followed by a multimodal MoE (Mixture of Experts) classifier.
The video-level features for video and audio are features respectively extracted from the visual and auditory stream over the length of the video,
and processed into a single feature map by the competition organizers.
The frame-level features are computed from one frame every second of the video, with audio features computed over the same window.
All features are computed using a truncated state-of-the-art deep learning classification model. Video-level features are computed over these by pooling over them.
Further details are given in the original paper~\cite{abu2016youtube}.

Next, we introduce how we build the YouTube-8M-Text dataset, and the three video-level text features built upon the text dataset.

\section{Learning Text Representations}
\label{sec:textmodels}

\subsection{YouTube-8M-Text Dataset}
\label{sec:dataset}

To ensure good data hygiene, only the video identifiers for the training and validation sets were released.
We use those identifiers to download associated text information, such as the video title and keywords.
The original training set is split into a new training set and a development set which we use to tune hyper-parameters.
The original validation set is used to evaluate the performance of text representations for video classification.
To preprocess keywords, we normalize case and remove punctuation, numbers, as well as two stop words, "and" and "the".
Similarly, titles are preprocessed by removing all symbols and punctuation only. Normalization of case and removal of stop words is not done so as to preserve sentence structure.
Subsequently, our pre-trained word embedding model does not include all non-English titles and keywords, thus they are discarded.

A Word2Vec model pre-trained on a Google News dataset, which is capable of capturing semantic regularities between vector embeddings of words \cite{word2vec1,word2vec2,word2vec3}, is then used to obtain 300 dimensional vector embeddings for each word.

In summary, we can only perform text analysis on about two-thirds of the dataset, due to different loss factors:
\begin{enumerate}
  \item the video may no longer be available online;
  \item external video embedding may be disabled (we used the external embedding API to download video metadata);
  \item data is in a non-English language; or
  \item the pre-trained Word2Vec model, which contains vectors for 3 million words and phrases, does not provide an embedding for some words.
\end{enumerate}

\begin{table}[h]
\begin{tabularx}{\columnwidth}{rcccc}\toprule
	 & \textbf{Train.}  & \textbf{Dev.} &  \textbf{Val.} & \textbf{Test} \\
	\midrule
	Video \& Audio ($\times 10^{6}$)& 3.905 & 1.000  & 1.401 & 0.701 \\
	\hline
	With keywords  ($\times 10^{6}$)& 2.578 & 0.659 & 0.921 & - \\	
	With titles ($\times 10^{6}$)& 3.084 & 0.790 & 1.103 & - \\	
	With both ($\times 10^{6}$)& 2.473 & 0.633 & 0.884 & - \\	
	\bottomrule
\end{tabularx}
\caption{Sizes of the YouTube-8M video and YouTube-8M-Text datasets. ``With'' keywords and title also means that they are in English (at least partly for keywords). Note that text data is not available for Test set as test video identifiers are not released. For training text models, we split the original Kaggle training dataset into a new training set and a development set,
on which we tune parameters.}
\label{tab:datasets}
\end{table}

Table \ref{tab:datasets} displays the population of all the different datasets.
Titles are generally short, descriptive phrases; while keywords are a collection of words relevant to the video without any particular order.
There are about 48,000 and 45,000 unique English keywords and title words respectively in the dataset.

\begin{figure}[!h]
        \includegraphics[width=\columnwidth]{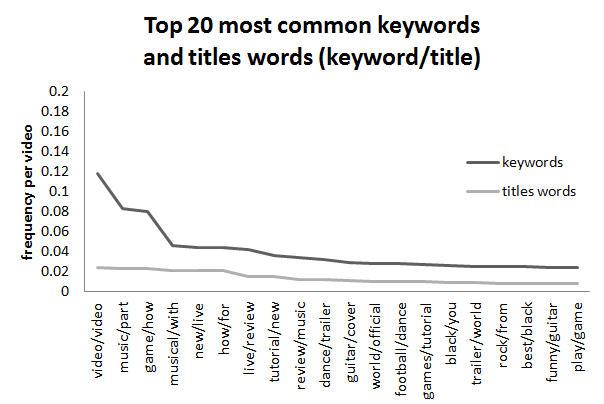}
        \caption{Keywords and title words distribution, showing the top 20 most common words from each set of words. In the x-axis words are shown in the keyword/title order. Note that some words are in common and at the same frequency rank, such as the most  common one ('video').}
        \label{fig:distribkw}
\end{figure}

\begin{figure}[!h]
        \includegraphics[width=\columnwidth]{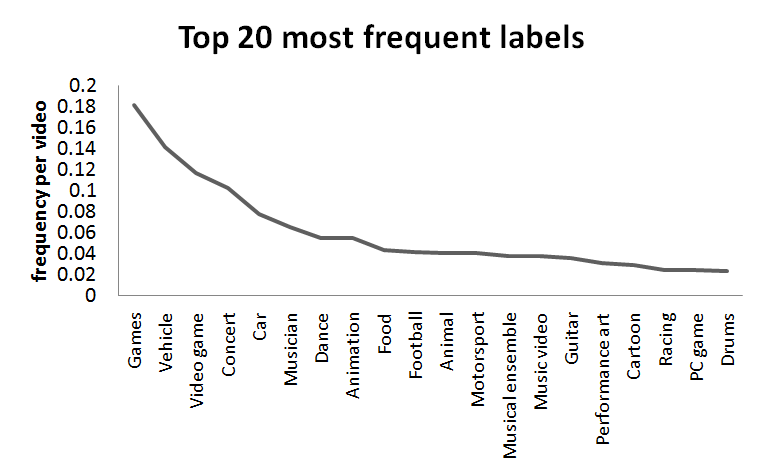}
        \caption{Label class distributions, showing the top 20 most common words. }
        \label{fig:distriblabel}
\end{figure}

A concern with using titles and keywords for YouTube8M challenge is that they may directly contain the class labels,
which could allow trained models to perform well without learning high-quality video and audio features.
Figure~\ref{fig:distribkw} and Figure~\ref{fig:distriblabel} show the top 20 most common words in both the keywords and the class labels sets. A few common concepts such as "game" and "dance" are inevitably highly ranked in both distributions. We discuss this issue in the experimental section.

\subsection{Unigram Model}

\begin{figure*}[t]
   \centering
   \includegraphics[width=\textwidth,trim={2.5cm 3cm 4cm 4.5cm},clip]{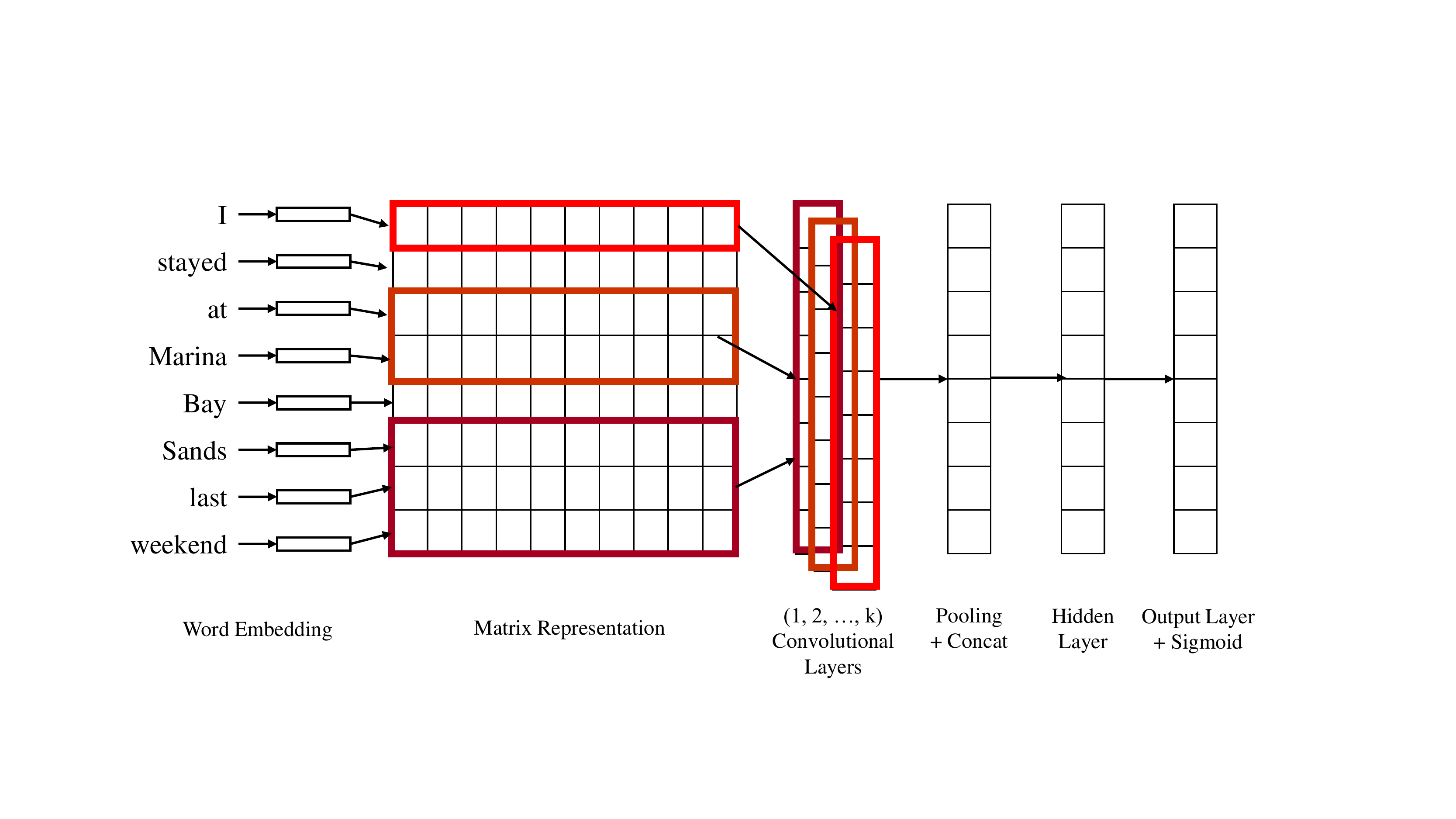}
   \caption{Architecture for the TextCNN model.}
   \label{fig:textcnn}
\end{figure*}

We use the classical unigram language model as the baseline against which to compare the text models introduced later.
In particular, given a sequence of words $k_1, k_2, \ldots, k_n$ from a title or set of keywords,
we assume that each word was drawn independently at random from a distribution $K_l$ that depends only on the label $l \in L$.
$L$ is the set of all labels, and $|L| = 4716$ in our dataset. We use the approximation:
\begin{align}
   & \Pr(L = l | k_1, k_2, \ldots, k_m)  \nonumber
\\ & \propto \Pr(k_1, k_2, \ldots, k_n | L = l) * \Pr(L = l)
\\ & = \Pr(L = l) * \prod_{i=1}^{m} \Pr(K_l = k_i | L = l)
\\ & \approx \frac{n(L = l)}{n(\circ)} * \prod_{i=1}^{m} \frac{n(K_l = k_i \cap L = l) + \alpha}{n(L = l) + \alpha|L|}
\end{align}

Where $n(\ldots)$ is the number of times such an event is observed in the training data.
We use $n(\circ)$ to indicate the total number of examples in the dataset.
We use Laplacian (add-alpha) smoothing (with $\alpha = 0.001$ - so that$\alpha|L|$ has an order of magnitude of 1) to allow our model to gracefully handle rare words. We test this model on 20,000 videos randomly chosen from the validation set.

\subsection{Histogram for Keywords}

As the order of keywords is assumed not to matter, we use a bag-of-words histogram approach.
Word2Vec vectors from all words in the training set are clustered into 1024 separate clusters, which are discovered through k-means clustering.
Each keyword is matched to the closest centroid, and this matching is expressed as a 1-hot vector.
The sum of these vectors is then the keyword feature.
This keyword feature is provided to a keyword classification decoder, and also one input of the MoE model (Fig.~\ref{fig:architecture_kaggle}).
The classification decoder is a neural network with two fully-connected layers followed by a sigmoid layer. Given the vast amount of data available, we trained this network on a quarter of the training data only, as it was not degrading performance and saving computation time.

Manual inspection of the training set reveals that it is not uncommon for YouTube users to, despite instructions to the contrary, attempt to include sets of words (or even phrases) in keywords.
This contradicts our assumption that the order of keywords does not matter.
Further work is necessary to establish if this effect is significant, and if models that account for word order would be better suited to parsing this data.

\subsection{TextCNN for Titles}

As the order of words in video titles are important, we train a multi-window convolutional neural network similar to \cite{journals/corr/Kim14f} that can capture individual semantic meanings as well as long-range dependencies in sentences.
A diagram of this architecture can be seen in Figure~\ref{fig:textcnn}.

After preprocessing, the pre-trained Word2Vec model is used to convert each word into a vector space $R^d$ where $d=300$, transforming a title with $n$ words into a $(n\times d)$ feature map, which is then zero padded to the maximum sentence length.
We then apply multiple 1d convolutions of different filter widths ($1, 2, ..., k$) in parallel to obtain feature maps for each filter width.
These feature maps are max pooled along the temporal dimension resulting in $k$ different $512$ dimensional vectors. We use $k=8$ (number found by cross-validation) for all experiments.
We concatenate them and use one hidden layer with 4096 neurons followed by an output layer with sigmoid activation function to predict each class.
Batch normalization is used before ReLU non-linearities to improve convergence. We train for 5 epochs with a batch-size of 512.
To regularize the model, we apply dropout on the second last layer as well as use an l2 weight decay of 1e-7.

Features from the second last layer of the network are also extracted and used as 4096-dimensional text features for the MoE model shown in Fig.~\ref{fig:architecture_kaggle}.

\section{Experiments}

\subsection{Evaluation Metrics}
We use four metrics to quantify the performance of their models.
The Kaggle YouTube8M competition focuses on the GAP metric, so we will primarily be reporting on GAP, but we will also include the other metrics for certain analyses.
These metrics are:

\begin{enumerate}
  \item Global Average Precision (GAP): the precision of a model from predictions over the entire test set.
  \item Mean Average Precision (mAP): the unweighted mean of area under the precision-recall curve for each class.
  \item Hit@$1$: 1 if at least one of the ground truth labels in the top prediction, 0 otherwise.
  \item Precision at equal recall rate (PERR): the precision when comparing the $j$ ground-truth labels for a video against the top $j$ predictions of our model.
\end{enumerate}

The mean value over the entire test set is reported for mAP, Hit@1, and PERR scores.
GAP scores are calculated over the test set, and so no averaging is necessary.

\subsection{Results on Kaggle Competition}

In this section, we report video classification results on YouTube-8M \textbf{Test} dataset, which ranked 22th out of 650 teams.
Besides Mixture of Expert (MoE) model with video-level visual and audio features,
we also investigate the use of frame-level models, i.e. Long Short Term Memory (LSTM) and Deep Bag of Frame (DBOF) with frame-level visual and audio features.
More technical details on MoE, LSTM and DBOF can be found in the original paper~\cite{abu2016youtube}.

\begin{table*}[t]
\centering

\begin{tabular}{| c | c | c | c|}
\hline
\multicolumn{2}{|c|}{Model} & Parameter Setting & GAP \\
\hline
\hline
\multirow{2}{*}{Video Level} & MoE A & $Experts=8$ & 79.44 \\
& MoE B & $Expert=32$ & \textbf{79.76} \\
\hline
\multirow{4}{*}{Frame Level} & DBOF A &	$Projection=8K, Hidden=1K,Experts=2, Samples=30$ & 79.27 \\
& DBOF B &	$Projection=16K, Hidden=2K, Experts=16, Samples=60$ &	\textbf{80.18} \\
& LSTM A &	$Cells=1K, Layers=2, Experts=2$ & \textbf{80.22} \\
& LSTM B &	$Cells=512, Layers=4, Experts=4$ & 79.30 \\	
\hline
\hline
\multicolumn{2}{|c|}{\multirow{3}{*}{Model Ensembling}} & Average Pooling	& \textbf{82.74} \\
\multicolumn{2}{|c|}{} & Max Pooling	& 81.56 \\
\multicolumn{2}{|c|}{} & Random forest	& 78.35 \\

\hline
\end{tabular}
\caption{Results of video-level and frame-level models on YouTube-8M Test set for Kaggle competition, with the concatenation of visual and audio features as input for each model. In this table, 'Projection' denotes the dimensionality of projection layer, 'Hidden' denotes the dimensionality of hidden layer, 'Experts' denotes the number of experts in MOE classifier, 'Samples' denotes sampling number for frame features, 'Cell' denotes number of parameter in one LSTM cell, and 'Layers' denotes the number of layer in LSTM.}
\label{table:kaggle}
\end{table*}

\begin{table}[t]
\centering

\begin{tabular}{| c | c | c | c |}
\hline
\multirow{2}{*}{Model} & \multicolumn{2}{c|}{Feature} & \multirow{2}{*}{GAP} \\
\cline{2-3}
& Visual & Audio & \\
\hline
\multirow{3}{*}{Mixture of Expert} & \tick & & 74.17 \\
& & \tick & 45.17 \\
& \tick & \tick & \textbf{78.18} \\
\hline
\end{tabular}
\caption{Results of visual feature and audio features with mixture-of-expert (MOE) model on YouTube-8M Test set for Kaggle competition.}
\label{table:visual_audio}
\end{table}

\begin{table}[t]
\centering

\begin{tabular}{| c | c | c |}
\hline
Rank	& \parbox[c]{4cm}{ Team Name} &	GAP \\
\hline
1 &	WILLOW &	84.97 \\
\hline
2 &	monkeytyping &	84.59 \\
\hline
3 &	offline &	84.54 \\
\hline
4 &	FDT &	84.19 \\
\hline
5 &	You8M &	84.18 \\
\hline
10 &	Samaritan &	83.66 \\
\hline
20 &	DeeepVideo &	82.91 \\
\hline
\textcolor{red}{22} &	\textcolor{red}{DL2.0 (Ours)} &	\textcolor{red}{82.74} \\
\hline
50 &	n01z3 &	81.50 \\
\hline
100 &	lwei &	80.37 \\
\hline
\end{tabular}
\caption{Leadboard of Kaggle competition.}
\label{table:ranking}
\end{table}

\textbf{Visual vs. Audio}.
It is not surprising that visual features are more discriminative than audio features.
As shown in Table~\ref{table:visual_audio}, with MoE model, visual features achieve $74.17\%$ GAP while audio features only $45.17\%$.
However, the visual and audio features are complementary to each other.
By simply concatenating video-level visual and audio features as the input to MoE model, there is a significant improvement in performance.
For example, MoE with visual and audio features is $4\%$ better than MoE with visual features only.
The same trend applies to frame-level models (LSTM and DBOF) as well.
For the following sections, we use the concatenation of visual and audio features.

We expect audio features to contribute more if feature extraction were conducted on smaller time frames; the default setting for audio features were set at 1 second buckets, which are too long for many audio, speech, or musical expressions. We will further investigate this in future work.

\textbf{Frame-level Model vs. Video-level Model}.
Table \ref{table:kaggle} shows results of frame level and video level models on YouTube-8M Test set.
For each model, we report two groups of results with different parameter settings.
Overall, frame level models DBOF and LSTM performance are better than video level model MoE.
DBOF and LSTM achieve $80.22\%$ and $80.18\%$ in terms of GAP, compared to MoE with $79.76\%$.
Both DBOF and MoE benefit from the increased parameters, while LSTM does not.
For example, by increasing dimensionality of projection layer, dimensionality of hidden layer, number of experts in classifier and sampling number,
DBOF obtains a considerable improvement with $+0.9\%$.
For LSTM, GAP drops from $80.22\%$ to $79.3\%$ with more experts and memory cells.

\textbf{Ensembling}.
Ensembling is an efficient way to boost performance.
Here, we test three model ensembling strategies:
\begin{enumerate}
 \item \textbf{Max pooling}, where the score for each class is the maximum value across all classifiers;
 \item \textbf{Average pooling}, where the score for each class is the mean of the score assigned by all classifiers;
 \item \textbf{Random forest}, where we construct 1000 trees and select the mode of the generated decisions.
\end{enumerate}
As shown in Table~\ref{table:kaggle}, average pooling strategy performs best than the other two ensembling methods.
By using average of our 6 independent models, we obtain $82.74\%$ in terms of GAP, which is our final number for the Kaggle competition.

\textbf{Leaderboard on Kaggle}.
Table~\ref{table:ranking} shows the rankings of YouTube-8M competition, we ranked 22th out of 650 teams.
This promising performance validates the effectiveness of combining video and audio models.

\begin{table*}[t]
\centering
\begin{tabular}{| c | c  c  c  c | c | c | c | c |}
\hline
\multirow{2}{*}{Model} & \multicolumn{4}{c}{Feature} & \multicolumn{4}{c|}{Performance} \\
\cline{2-9}
& Visual &	Audio &	Keyword &	Title &	GAP &	Hit@1 &	PERR &	mAP \\
\hline
\hline
\multirow{5}{*}{Mixture of Expert} & \tick &	\tick & &	&		\textbf{77.6} &	83.7 &	70.2 &	31.9 \\
& 		\tick & & & 	&	35.8 &	46.4 &	35.9 &	14.5 \\
& 		&	\tick & & 	& 60.7 &	71 &	57.8 &	49.5 \\
& 		& & \tick & \tick & 65.4 &	75.8 &	62.0 &	50.0 \\
& \tick &	\tick &	\tick & &		79.6 &	85.9 &	72.8 & 	42.9 \\
& \tick &	\tick &	\tick &	\tick &	\textcolor{red}{\textbf{86.7}} &	91.5 &	80.4 &	62.8 \\
\hline
\hline
Model on Kaggle & 	\tick &	\tick & &	&		\textbf{81.1} &	- &	- &	-\\
\hline
\end{tabular}
\caption{Results of the proposed multi-modal learning framework with visual, audio and text features, on YouTube-8M-Text Validation set.
Note that the last line also reports result on the same validation set, using our ensembling model for Kaggle competition.}
\label{table:text}
\end{table*}

\begin{table}
\centering

\begin{tabular}{| c | c | c | c |}
\hline
\multirow{2}{*}{Model} & \multicolumn{2}{c|}{Feature} & GAP \\
\cline{2-4}
 & Title & Keyword & \\
 \hline
 TextCNN (Sigmoid) A &	\tick	&  &	\textbf{53.50} \\
TextCNN (Sigmoid) B &	\tick & &		41.00 \\
\hline
\hline
Histogram (Sigmoid) A & & \tick &	\textbf{46.20} \\
Histogram (Sigmoid) B	& &	\tick &	38.90 \\
\hline
\hline
Unigram & \tick &  &		19.2 \\
Unigram & & \tick &		22.60 \\
\hline
\end{tabular}
\caption{Performance of text models with different parameters when evaluating on YouTube-8M-Text Validation set.
Both TextCNN model A and B set parameters Filter Widths=(1, 2, ..., 8) and Filter Channels=512.
TextCNN model B uses filter labels. Histogram B also uses filter labels but Histogram A model doesn't. }
\label{tab:params_text}
\end{table}

\subsection{Results with Text Features}

In this section, we evaluate the performance of text incorporated with video and audio features for a truly multi-modal video classification.
As text information is not available for YouTube-8M Test set, we report all results on YouTube-8M-Text Validation set instead (see Table \ref{tab:datasets}).

\subsubsection{Incorporating Text Features}

We perform experiments to quantify the performance of visual, audio, and text features with MoE model.
Table~\ref{table:text} presents the results, we compare the performance of MoE model with and without text features.
We observe that MoE with visual and audio features alone can achieve 77.6\% GAP.
The full model with visual, audio and text features obtains 86.7\% GAP, which is significantly better.

It's worth mentioning that for the Kaggle competition our best model ensembling approach with visual and audio features obtained $82.7\%$ on YouTube-8M Test set,
while it performs a bit worse on the YouTube-8M-Text Validation set, i.e. $81.1\%$.
We think the reason why our model in Kaggle competition performs slightly worse on Validation set is that Validation set contains more testing videos than Kaggle test dataset (0.88M vs 0.7M).
Compared with our model on Kaggle competition with only visual and audio features, the MoE model with visual, audio and text features achieves significantly better performance (86.7\% vs. 81.1\%) leading to an 8.9\% absolute gain.

\begin{figure*}[t]
        \centering
        \includegraphics[width=0.8\textwidth]{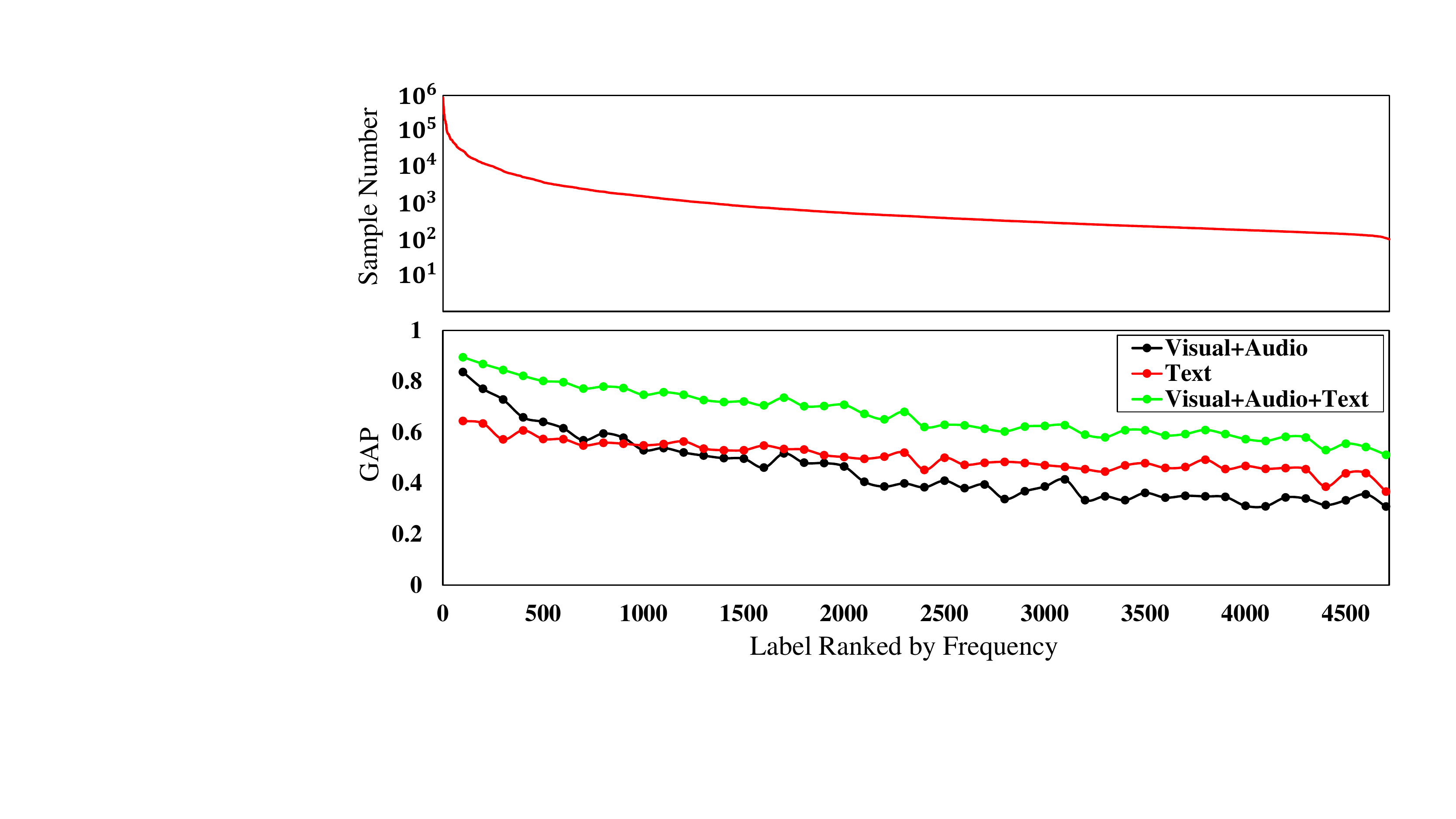}
        \caption{Performance of visual, audio and text features across labels. Labels (x-axis) are ranked by frequency.}\label{fig:ana}
\end{figure*}

\subsubsection{Discussions}

\textbf{How does text help audio and video ?}
In order to investigate if some labels are more easily classified using certain modes, we compute the GAP for each label using video and audio, text, and all three features.
We sort all labels based on their frequency and then graph the GAP against the label rankings.
As seen in Fig.~\ref{fig:ana}, this allows us to examine the performance over classes and across features.

We observe two key points in Fig.~\ref{fig:ana}: First, there is a severe imbalance in the number of videos available for each label: the top 10 labels account for just over half of all label appearances.
Second, while the MoE model with video and audio features dominates at first,
there is a long tail where the MoE model with text features is better than the model with video and audio features.
This suggests that the text mode helps a lot when the visual and audio modes together are unable to recognize a label with few training samples.
This is probably due to that text feature contains relatively high-level semantic information, thus less sensitive to the amount of training data (red curve - Text is flatter than black curve - Visual + Audio).
In this sense, text is complementary to video and audio, leading to better performance when combining all of them.

\textbf{Unigram, Histogram, TextCNN}.
We present the evaluation of our text models in Table~\ref{tab:params_text}, with an additional baseline for comparison.
The \emph{unigram} baseline reports the expected performance if we were to use a simple Bayesian model to predict the label from text input.
This performance is calculated on validation set videos that include at least one word that is recognized by the pre-trained Word2Vec model.

One concern is that the ground truth label appears among the keywords or titles frequently enough for the text models to trivially predict labels.
We observe that this is true only in about 10\% of examples with title or keyword data. To get this overlap for labels containing several words, we establish that such a label is present in the keywords or the title if all its words are. We get a ratio of keywords or titles words that are among the video's labels, and then average this ratio on all videos.

The unigram baseline model is only slightly better than the overlap, providing a GAP of 19.2\% and 22.6\% for titles and keywords respectively. In comparison, the keywords histogram model with classifier achieves a GAP of 46.2\% while the TextCNN achieves a GAP of 53.5\%.
Moreover, fusion of features from both models with a Mixture of Experts achieves 65.4\% GAP, suggesting that there is a deep, non-linear relationship in the text that the unigram model is unable to capture.

\textbf{Titles and Keywords with or without Labels}.
As mentioned in the previous subsection, we observe a roughly 10\% overlap between the labels and both the keywords and titles. It does not constitute such a surprise given that keywords, titles and labels are three sets of words of approximatively the same length describing the same video.
This also suggests that the keywords, despite the risk of machine suggestion bias, are not really more predictive of the label than the purely human-generated title. Thus, they contain non-biased information that can help the video classification.

Further, we show that our method generalizes well by deliberately changing the distribution of the data to minimize any effect the bias may have.
We retrain our model to deliberately exclude all words from keywords and titles that match that video's label.
Under these conditions, a GAP of 38.9\% is observed for the keywords classifier, and a GAP of 41.0\% is observed when retraining the TextCNN.

The loss in GAP of only 7.3\% for keywords and 12.5\% for titles of performance in Table~\ref{tab:params_text} shows that our system has learned good features from the video, audio, and title text, and is not reliant on trivial matching of keyword labels, as we are still consequently above the unigram performance. A larger GAP loss for titles than for keywords could be explained by the fact that word order matters in titles. Removing words breaks sentences, thus harming training more than in keywords.

\textbf{Predicting on learned predictions}.
We observe that the highest mAP score on the Kaggle competition was about 85\%, with over thirty teams achieving 83\%.
The state-of-the-art mAP of the next largest video classification task--ActivityNet--is comparatively poor at 77.6\% when model training is assisted with the YouTube-8M dataset \cite{abu2016youtube}.
ActivityNet \cite{caba2015activitynet} is a comparatively easier task, with only 203 classes.

We believe that this disparity can be explained by a strong bias in selecting videos for this dataset. In their paper introducing the challenge,
\cite{abu2016youtube} explain that the videos are machine-generated and selected for precision
\footnote{``While the labels are machine-generated, they have high-precision and are derived from a variety of human-based signals...'' \cite[p.~1]{abu2016youtube}},
before being subject to additional filtering to improve precision.
The dataset, therefore, is strongly biased towards videos that can be classified with high precision based on associated ``human-based signals''.
If we make the reasonable assumption that videos with clearer visual and auditory content also have clearer human-based signals,
then it follows that videos from the dataset will be easier to classify than the average YouTube videos.
Further, if these signals \emph{directly} depend on video or audio data, the task reduces to us attempting to learn another machine learning model.

On a sample of videos randomly selected over YouTube, we would expect classification performance to be substantially worse.

\section{Conclusion and Future Work}

Video classification has long been an interesting open problem with tremendous potential applications.
Video classification approaches so far have only used the raw video and audio as input, eschewing the vast amounts of metadata available.
Our work makes this classification truly multimodal by incorporating video titles and keywords,
and showing a tremendous improvement over state-of-the-art competition models on the world's largest video classification dataset.

The YouTube-8M team provided audio features in a later release \cite{yt8mAudioRelease}, where they indicated that audio features were processed in 1-second buckets.
The procedure that was followed to extract these features is comparable to their image extraction, but is not particularly suited to audio information:
the window is too large to allow for extraction of short-lived sounds, such as musical notes, explosions, word utterances, etc.
This could be greatly improved by processing raw audio data directly.
Given the relative complexity of audio and visual data, this could be achieved without substantially increasing the computation required to train the multimodal model.

We are investigating the use of the official YouTube metadata API to download the data that is available but for which embedding is disabled,
which should increase the number of videos for which text data is available.
We are also considering approaches to generate word embeddings for non-English languages, and integrating them into our pipeline.

Human annotations are expensive, and so machine-assisted annotations were likely the most cost-effective way to assemble this large dataset.
However, as we have discussed, we are likely training and testing on data that is \emph{a priori} more amenable to machine classification.
The only obvious mechanism to eliminate this bias is to collect high-quality annotations for randomly selected YouTube videos instead of selecting videos based on annotation quality;
a solution that is unpalatably expensive. Perhaps we could investigate the strength of this bias with access to examples annotated with the confidence that the automated rating system assigns to them.
If the YouTube-8M team were to release such data,
it would allow us to evaluate models on high- and low-confidence data and allow us to gain a more realistic evaluation of the real-world performance of our models.

Also, our multimodal MoE model does not take advantage of the performance gains afforded by ensembling with diverse models.
We need to devise additional models that are multimodal and ensemble them to further improve our classifier performance.

Finally, we can apply the intermediate representations learned in this problem to other video tasks, such as video retrieval.

\section{Acknowledgments}

This work was supported by the SUTD-MIT International Design Centre Grant IDIN16007, SUTD grant SRG ESD 2015 097 and MOE AcRF Tier 2 Grant 2016-T2-1-170.

{
\bibliographystyle{ieee}
\bibliography{egbib}

\begin{thebibliography}{1}\itemsep=-1pt

\bibitem{abu2016youtube}
S.~Abu-El-Haija, N.~Kothari, J.~Lee, P.~Natsev, G.~Toderici, B.~Varadarajan,
  and S.~Vijayanarasimhan.
\newblock Youtube-8m: A large-scale video classification benchmark.
\newblock {\em arXiv preprint arXiv:1609.08675}, 2016.

\bibitem{caba2015activitynet}
F.~Caba~Heilbron, V.~Escorcia, B.~Ghanem, and J.~Carlos~Niebles.
\newblock Activitynet: A large-scale video benchmark for human activity
  understanding.
\newblock In {\em Proceedings of the IEEE Conference on Computer Vision and
  Pattern Recognition}, pages 961--970, 2015.

\bibitem{yt8mAudioRelease}
S.~H. et.al.
\newblock Cnn architectures for large-scale audio classification, ICCASP, 2017.

\bibitem{word2vec3}
Google.
\newblock word2vec.

\bibitem{karpathy2014large}
A.~Karpathy, G.~Toderici, S.~Shetty, T.~Leung, R.~Sukthankar, and L.~Fei-Fei.
\newblock Large-scale video classification with convolutional neural networks.
\newblock In {\em Proceedings of the IEEE conference on Computer Vision and
  Pattern Recognition}, pages 1725--1732, 2014.

\bibitem{journals/corr/Kim14f}
Y.~Kim.
\newblock Convolutional neural networks for sentence classification.
\newblock {\em CoRR}, abs/1408.5882, 2014.

\bibitem{word2vec1}
T.~Mikolov, K.~Chen, G.~Corrado, and J.~Dean.
\newblock Efficient estimation of word representations in vector space.
\newblock {\em arXiv preprint arXiv:1301.3781}, 2013.

\bibitem{word2vec2}
T.~Mikolov, W.-t. Yih, and G.~Zweig.
\newblock Linguistic regularities in continuous space word representations.
\newblock In {\em Hlt-naacl}, volume~13, pages 746--751, 2013.

\end{thebibliography}
}

\end{document}